%% file: iclr2025_conference.tex
\documentclass{article} 
\usepackage{iclr2025_conference,times}
\usepackage{booktabs}
\input{math_commands.tex}

\usepackage{pgfplots}
\usepackage{pgfplotstable}
\usepackage{tikz}
\usepackage{hyperref}
\usepackage{url}
\usepackage{wrapfig}
\usepackage{graphicx} 

\begin{document}

\begin{center}
    \includegraphics[width=0.2\textwidth]{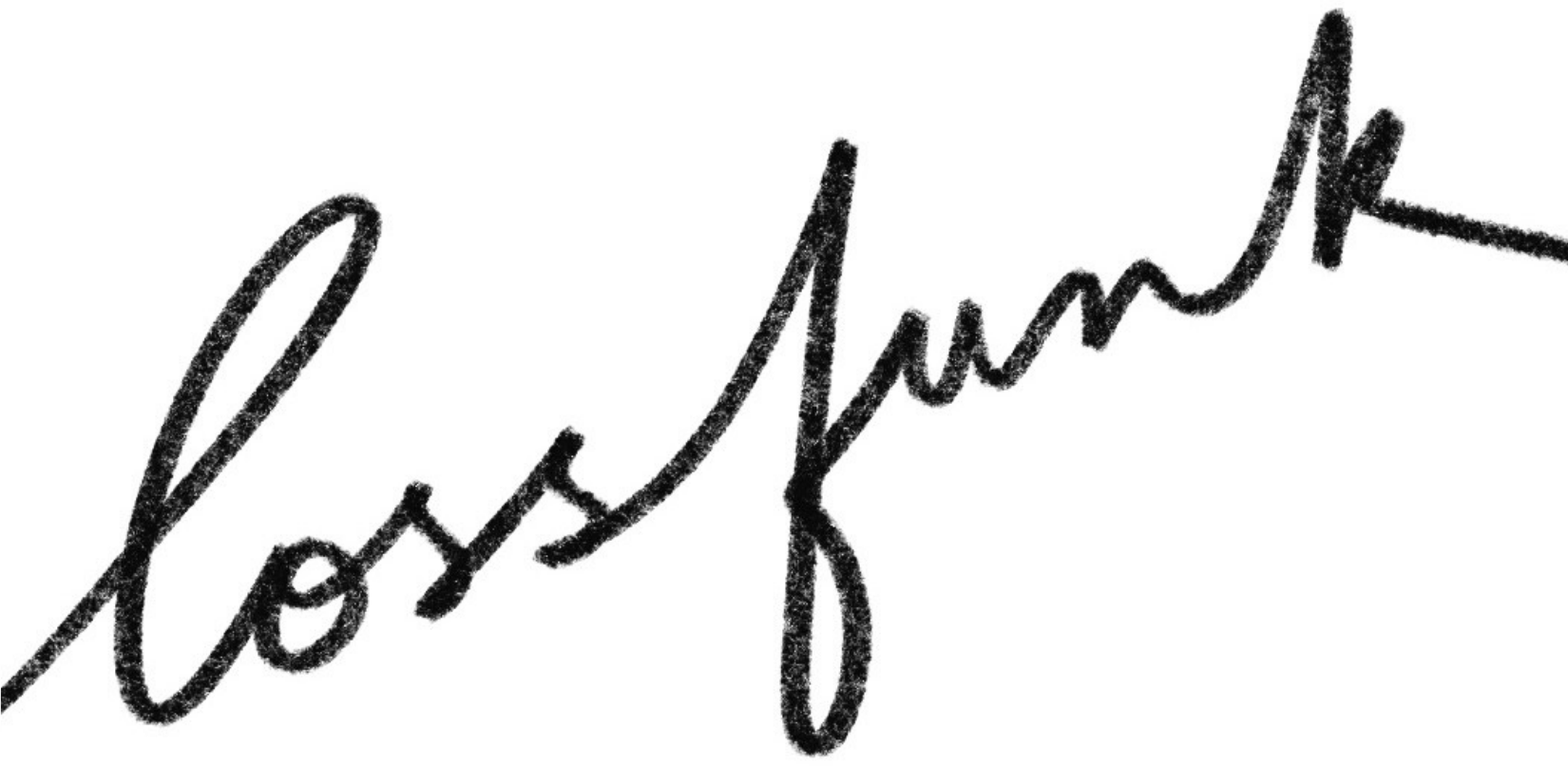} 
\end{center}

\title{Do GFlowNets Transfer? Case Study on the Game of 24 / 42}

\iclrfinalcopy
\author{Adesh Gupta, Abhinav Kumar, Mansi Gupta \\
Indian Institute of Technology, Roorkee\\
\texttt{\{adesh\_g, abhinav\_k, m\_gupta\}@ma.iitr.ac.in} \\
\And
Paras Chopra \\
Lossfunk \\
\texttt{paras@lossfunk.com} \\
}

%

\newcommand{\fix}{\marginpar{FIX}}
\newcommand{\new}{\marginpar{NEW}}


\maketitle

\begin{abstract}

Generating diverse solutions is key to human-like reasoning, yet autoregressive language models focus on single accurate responses, limiting creativity. GFlowNets optimize solution generation as a flow network, promising greater diversity. Our case study shows their limited zero-shot transferability by fine-tuning small and medium-sized large language models on the Game of 24 and testing them on the Game of 42 datasets. Results revealed that GFlowNets struggle to maintain solution diversity and accuracy, highlighting key limitations in their cross-task generalization and the need for future research in improved transfer learning capabilities.

\end{abstract}

\section{Introduction}

Recent advances have introduced approaches showing significant improvement in LLM reasoning capabilities \citep{touvron2023llama}, including supervised fine-tuning with synthetic datasets \citep{yumetamath, yuemammoth}, modified decoding mechanisms \citep{holtzmancurious, nguyen2024turning}, and enhanced pretraining data quality \citep{akter2024mind, trinh2024solving}. While these approaches demonstrate improved accuracy, they rarely account for the diversity of correct solutions, an essential aspect of human-like reasoning and creativity \citep{yu2024flow, huamortizing}.
GFlowNets \citep{bengio2023gflownet,bengio2021flow} tackle this problem by learning from probability distributions proportional to reward functions, enabling diverse, high-reward sequence generation \citep{yu2024flow, huamortizing}. However, systematic empirical evaluations of these methods remain limited. This case study takes a step in that direction by examining GFlowNets' transfer learning ability on the closely related Game of 24 and 42 datasets. While GFlowNets excel at generating diverse, valid solutions for in-distribution problems, their performance drops significantly on out-of-distribution tasks. Our analysis reveals that their generalization heavily depends on hyperparameter configurations, in contrast to the robust zero-shot transfer capabilities observed in instruction-tuned models(\citep{weifinetuned, wang2023self, wang2022super}).

\section{Background and Related work}

\subsection{Decoding Strategies and Temperature}
We experiment with the following different decoding strategies:
\begin{itemize}

    \item \textbf{Top-k Sampling:} Samples from the top $k$ most probable tokens, renormalizing probabilities across this set \citep{fan2018hierarchical}.
    
    \item \textbf{Top-p Sampling:} Also known as nucleus sampling, it selects tokens whose cumulative probability exceeds a threshold $p$ \citep{holtzmancurious}. 
    
    \item \textbf{Min-p Sampling:} Filters out tokens below a minimum probability threshold $p_{min}$, sampling is done from the remaining set\citep{nguyen2024turning}.
\end{itemize}

Temperature scaling adjusts the sharpness of probability distributions, balancing diversity and coherence in language model decoding \citep{ackley1985learning}. Lower temperatures (0.2–0.5) concentrate probability mass on high-likelihood tokens, enhancing precision for tasks like mathematical reasoning. In contrast, higher temperatures (1.0–1.5) flatten the distribution, promoting diversity but potentially reducing coherence and logical consistency.

\subsection{GflowNets}

GFlowNets \citep{bengio2021flow} were introduced to generate diverse, high-reward samples from unnormalized distributions, contrasting with traditional reinforcement learning approaches \citep{schulman2017proximal} that focus solely on reward maximization. Recent applications to LLM fine-tuning have shown promising results in tasks like story infilling and logical reasoning (\cite{yu2024flowreasoningtrainingllmsdivergent}, \cite{hu2024amortizingintractableinferencelarge}). Our work builds on these developments by investigating GFlowNets' potential for generating correct alternative solutions, evaluating their fine-tuning benefits regarding solution accuracy and diversity, and examining zero-shot transfer learning capabilities across related domains.

\section{Game of 24 and Game of 42}

The Game of 24 is a mathematical reasoning challenge \citep{yao2024tree}, where the objective is to use four given numbers and basic arithmetic operations (\(+, -, *, /\)) to achieve the target value of 24. Extending this concept, we curated the Game of 42 dataset, where the goal is to achieve a target value of 42 using similar operations. Although 42 was chosen as the target for this study, other numbers could be selected to design similar reasoning tasks. The Game of 24 and its extension to the Game of 42 were chosen for their ability to test mathematical reasoning, problem-solving, and combinatorial logic in a structured yet diverse setting. These games challenge the model to apply arithmetic operations creatively while adapting to different target values. Their scalability and varying complexity make them ideal for systematically evaluating the accuracy and diversity of reasoning.
To systematically assess reasoning capabilities, we constructed two distinct subsets of problems for both games, designed to determine accuracy and diversity separately.

\begin{enumerate}
    \item \textbf{Subset 1: Low-Solution Diversity} \\
    In this subset, we included samples where the target value (24 or 42) can be achieved in only 1-2 unique ways. For example, the array \( (6, 6, 6, 6) \) can produce 24 in two ways:
       \( 6 \times (6 - 6) + 6 \) and 
       \( (6 + 6) \div (6 \div 6) \)
   
    This subset allows for a focused evaluation of accuracy, testing whether GFlowNets can reliably produce correct solutions for problems with limited answer spaces.
    
    \item \textbf{Subset 2: High-Solution Diversity} \\
    This subset includes samples where the target value can be achieved in at least seven distinct ways. For example, arrays such as \( (8, 3, 1, 6) \) can be combined in multiple ways to produce 24, ensuring a rich solution space. This subset is used to assess the diversity of solutions generated by GFlowNets, a key claim of their design.
\end{enumerate}

\section{Metrics}

To assess the performance of GFlowNet fine-tuned LLM on the Game of 24 and Game of 42, we employ two quantitative metrics, evaluating each problem instance over 20 sampling iterations.
\begin{itemize}
    \item \textbf{Success Rate (SR):} This binary metric indicates whether the model successfully generates a valid solution to achieve the target number (24 or 42) at least once within the 20 attempts allocated. This metric allows us to quantify the accuracy of the model.
    \item \textbf{Trajectory Count (TC):} This metric quantifies the number of unique and valid solutions generated by the model in the 20 attempts, with a minimum value of 0 and a maximum of 20. By definition, TC must be greater than or equal to SR. This metric allows us to quantify the ability of the model to generate diverse, unique, and correct solutions.
\end{itemize}

We average the results across a test set comprising 50 distinct problems to ensure comprehensive analysis. This comprehensive evaluation framework establishes a consistent foundation for assessing both in-distribution and out-of-distribution tasks.

\section{Experiments and Results}
We draw reference from \cite{yu2024flowreasoningtrainingllmsdivergent} and use the LLM-reasoner dataset (\cite{hao2024llmreasonersnewevaluation}) to select 20 examples (10 easy and 10 hard) for training, with examples ranked from easy to hard with easy ones having higher number of diverse solution as compare to harder ones. The success reward gives a high positive reward of 100 when a trajectory succeeds in
reaching 24 and a small reward of 0.001 otherwise. The dataset is used to train the model, which is then evaluated separately on both the Game of 24 and Game of 42 datasets, for different temperatures and decoding strategies. 

Previous work (e.g., \cite{yao2024tree}) highlights that LLMs struggle to sample correct trajectories. To address this, we sample each example 20 times, effectively allowing the model 20 attempts to arrive at the correct solution. Our experiments involve LLaMA-1B, LLaMA-3B, and LLaMA-8B (\cite{touvron2023llamaopenefficientfoundation}). The experiments are conducted in an eliminatory manner, starting with three values each for top-k, min-p, and top-p. We iteratively refine the search space to identify optimal parameters for larger models (8B) and diversity analysis experiments, conducting experiments on the best values. Also, it is important to note that the number of possible 4-tuples leading to a valid solution is greater for 42 than for 24, leading to inherently better performance on pre-trained models, in the case of Game of 42.

\subsection{Accuracy Analysis}
The results (Table \ref{tab:game241}) demonstrate that the fine-tuned GFlowNet consistently outperforms the standard pre-trained LLaMA models (\cite{touvron2023llamaopenefficientfoundation}) across all temperature settings and decoding strategies when evaluated on the Game of 24 dataset itself. Remarkably, the fine-tuned 1B LLaMA model surpasses the pre-trained LLaMA 8B model, highlighting the benefits of fine-tuning for task-specific performance. However, the model's ability to transfer knowledge to closely related tasks, such as the Game of 42, remains limited, as seen in Table  \ref{tab:game422}.  
Although certain temperatures combined with specific decoding strategies occasionally result in successful transfer, these instances are sparse and inconsistent.
Interestingly, we note that the maximum variation in performance occurs with temperature settings, highlighting the importance of temperature as a key hyperparameter. In contrast, we observe less variation with different decoding strategies. Top-k provides the most consistent results in the decoding strategy, with more stable performance across various configurations, and the optimal value for k is 10. By analyzing the findings, we determined the most optimal values for key parameters, such as 10 for top-k sampling, 0.05 for min-p sampling, and 0.85 for top-p sampling. We report the success rate (SR) value in these different settings. Complete results can be found in Appendix \ref{res} (refer Tables \ref{tab:game245} to \ref{tab:game4210}).

\subsection{Diversity Analysis}

To analyze the diversity transfer of GFlowNets, we use subsets of numbers with more than seven distinct solutions, checking an essential claim of GFlowNets over RL-based approaches. We used previously determined optimal values and evaluated the models while varying the temperature.
Our findings show a slight increase in diversity when evaluated on the Game of 24 datasets (refer Table \ref{tab:game243}). However, GFlowNets show no notable improvement in diversity for the Game of 42 datasets (refer Table \ref{tab:game424}). As with the accuracy analysis, we observe higher variation with temperature, with higher temperature settings(0.7, 1.1) yielding better performance by facilitating more exploration of possible solutions. We report the TC/SR metric, which intuitively represents the average number of unique correct solutions per correctly answered question (the diversity). A higher SR is naturally more desirable if the diversity value remains the same. To account for this, we also present the SR of all models across different settings in Appendix \ref{res} (refer Tables \ref{tab:game24-div} and \ref{tab:game42-div})
\begin{table*}[htbp]
\centering
\small
\begin{minipage}{0.48\textwidth}
\caption{SR of the models on GAME 24}
\vspace{5pt}
\centering
\begin{tabular}{lccc}
\hline
\textbf{Temperature} & \textbf{Top-10} & \textbf{Min-0.05} & \textbf{Top-0.85} \\
\hline
\multicolumn{4}{c}{\textbf{LLaMA 1B}} \\
\hline
0.3 & 0.04 & 0.04 & 0.02 \\
0.7 & 0.06 & 0.10 & 0.10 \\
1.1 & 0.04 & 0.06 & 0.10 \\
\hline
\multicolumn{4}{c}{\textbf{LLaMA 1B Fine-Tuned with GFlowNet}} \\
\hline
0.3 & 0.26 & 0.22 & 0.22 \\
0.7 & 0.30 & 0.26 & 0.24 \\
1.1 & 0.20 & 0.20 & 0.24 \\
\hline
\multicolumn{4}{c}{\textbf{LLaMA 3B}} \\
\hline
0.3 & 0.12 & 0.12 & 0.06 \\
0.7 & 0.22 & 0.24 & 0.22 \\
1.1 & 0.12 & 0.22 & 0.22 \\
\hline
\multicolumn{4}{c}{\textbf{LLaMA 3B Fine-Tuned with GFlowNet}} \\
\hline
0.3 & 0.36 & 0.36 & 0.30 \\
0.7 & 0.40 & 0.46 & 0.32 \\
1.1 & 0.30 & 0.30 & 0.26 \\
\hline
\multicolumn{4}{c}{\textbf{LLaMA 8B}} \\
\hline
0.3 & 0.24 & 0.32 & 0.24 \\
0.7 & 0.08 & 0.16 & 0.16 \\
1.1 & 0.16 & 0.12 & 0.40 \\
\hline
\multicolumn{4}{c}{\textbf{LLaMA 8B Fine-Tuned with GFlowNet}} \\
\hline
0.3 & 0.28 & 0.32 & 0.32 \\
0.7 & 0.52 & 0.28 & 0.40 \\
1.1 & 0.24 & 0.36 & 0.40 \\
\hline
\end{tabular}
\label{tab:game241}
\end{minipage}
\hfill
\begin{minipage}{0.48\textwidth}
\caption{SR of the models on GAME 42}
\vspace{5pt}
\centering
\begin{tabular}{lccc}
\hline
\textbf{Temperature} & \textbf{Top-10} & \textbf{Min-0.05} & \textbf{Top-0.85} \\
\hline
\multicolumn{4}{c}{\textbf{LLaMA 1B}} \\
\hline
0.3 & 0.10 & 0.12 & 0.10 \\
0.7 & 0.12 & 0.18 & 0.22 \\
1.1 & 0.12 & 0.08 & 0.14 \\
\hline
\multicolumn{4}{c}{\textbf{LLaMA 1B Fine-Tuned with GFlowNet}} \\
\hline
0.3 & 0.10 & 0.06 & 0.04 \\
0.7 & 0.26 & 0.20 & 0.14 \\
1.1 & 0.26 & 0.28 & 0.30 \\
\hline
\multicolumn{4}{c}{\textbf{LLaMA 3B}} \\
\hline
0.3 & 0.22 & 0.14 & 0.14 \\
0.7 & 0.16 & 0.28 & 0.18 \\
1.1 & 0.14 & 0.24 & 0.24 \\
\hline
\multicolumn{4}{c}{\textbf{LLaMA 3B Fine-Tuned with GFlowNet}} \\
\hline
0.3 & 0.16 & 0.10 & 0.04 \\
0.7 & 0.20 & 0.16 & 0.30 \\
1.1 & 0.24 & 0.12 & 0.30 \\
\hline
\multicolumn{4}{c}{\textbf{LLaMA 8B}} \\
\hline
0.3 & 0.24 & 0.32 & 0.36 \\
0.7 & 0.40 & 0.44 & 0.28 \\
1.1 & 0.28 & 0.36 & 0.44 \\
\hline
\multicolumn{4}{c}{\textbf{LLaMA 8B Fine-Tuned with GFlowNet}} \\
\hline
0.3 & 0.26 & 0.32 & 0.38 \\
0.7 & 0.44 & 0.36 & 0.26 \\
1.1 & 0.30 & 0.40 & 0.46 \\
\hline
\end{tabular}
\label{tab:game422}
\end{minipage}
\end{table*} 

\begin{table*}[htbp]
\centering
\small
\begin{minipage}{0.48\textwidth}
\caption{TC/SR of the models on GAME 24}
\vspace{5pt}
\centering
\begin{tabular}{lccc}
\hline
\textbf{Temperature} & \textbf{Top-10} & \textbf{Min-0.05} & \textbf{Top-0.85} \\
\hline
\multicolumn{4}{c}{\textbf{LLaMA 1B}} \\
\hline
0.3 & 1.5 & 1.5 & 2.0 \\
0.7 & 2.0 & 1.4 & 1.5 \\
1.1 & 1.24 & 1.2 & 1.0 \\
\hline
\multicolumn{4}{c}{\textbf{LLaMA 1B Fine-Tuned with GFlowNet}} \\
\hline
0.3 & 2.0 & 1.86 & 1.8 \\
0.7 & 3.3 & 1.88 & 2.68 \\
1.1 & 1.5 & 1.80 & 2.8 \\
\hline
\multicolumn{4}{c}{\textbf{LLaMA 3B}} \\
\hline
0.3 & 1.14 & 1.14 & 1.00 \\
0.7 & 1.00 & 1.17 & 2.20 \\
1.1 & 1.17 & 1.00 & 1.00 \\
\hline
\multicolumn{4}{c}{\textbf{LLaMA 3B Fine-Tuned with GFlowNet}} \\
\hline
0.3 & 1.5 & 2.2 & 2.00 \\
0.7 & 1.33 & 1.88 & 2.80 \\
1.1 & 1.00 & 2.86 & 1.5 \\

\hline
\multicolumn{4}{c}{\textbf{LLaMA 8B}} \\
\hline
0.3 & 1.08 & 1.44 & 1.00 \\
0.7 & 1.67 & 1.15 & 1.14 \\
1.1 & 1.14 & 1.27 & 1.09 \\
\hline
\multicolumn{4}{c}{\textbf{LLaMA 8B Fine-Tuned with GFlowNet}} \\
\hline
0.3 & 1.38 & 1.62 & 1.85 \\
0.7 & 1.76 & 1.85 & 1.26 \\
1.1 & 1.34 & 1.33 & 1.20 \\
\hline

\end{tabular}

\label{tab:game243}
\end{minipage}
\hfill
\begin{minipage}{0.48\textwidth}
\caption{TC/SR of the models on GAME 42}
\vspace{5pt}
\centering
\begin{tabular}{lccc}
\hline
\textbf{Temperature} & \textbf{Top-10} & \textbf{Min-0.05} & \textbf{Top-0.85} \\
\hline
\multicolumn{4}{c}{\textbf{LLaMA 1B}} \\
\hline
0.3 & 0.0 & 0.0 & 1.0 \\
0.7 & 1.0 & 1.0 & 1.0 \\
1.1 & 1.0 & 1.0 & 1.0 \\
\hline
\multicolumn{4}{c}{\textbf{LLaMA 1B Fine-Tuned with GFlowNet}} \\
\hline
0.3 & 0.0 & 1.0 & 0.0 \\
0.7 & 1.0 & 1.0 & 1.0 \\
1.1 & 1.0 & 1.0 & 1.0 \\
\hline
\multicolumn{4}{c}{\textbf{LLaMA 3B}} \\
\hline
0.3 & 1.00 & 1.00 & 1.33 \\
0.7 & 1.12 & 1.00 & 1.12 \\
1.1 & 1.13 & 1.20 & 1.62 \\
\hline
\multicolumn{4}{c}{\textbf{LLaMA 3B Fine-Tuned with GFlowNet}} \\
\hline
0.3 & 1.12 & 1.00 & 1.5 \\
0.7 & 1.30 & 1.00 & 1.0 \\
1.1 & 1.00 & 1.25 & 1.0 \\
\hline
\multicolumn{4}{c}{\textbf{LLaMA 8B}} \\
\hline
0.3 & 1.51 & 2.0 & 1.0 \\
0.7 & 1.53 & 1.76 & 1.3 \\
1.1 & 1.48 & 1.43 & 2.0\\
\hline
\multicolumn{4}{c}{\textbf{LLaMA 8B Fine-Tuned with GFlowNet}} \\
\hline
0.3 & 1.67 & 1.5 & 1.24 \\
0.7 & 1.55 & 1.81 & 1.35 \\
1.1 & 1.83 & 1.47 & 1.71 \\
\hline
\end{tabular}

\label{tab:game424}
\end{minipage}
\end{table*}

\section{Conclusion}
\pgfplotsset{compat=1.18}

\begin{wrapfigure}{r}{0.40\textwidth} 
     \vspace{-10pt}
     
    \centering

    \begin{tikzpicture}
        \begin{axis}[
            xlabel={Temperature},
            ylabel={Top-10 Score},
            xmin=0.3, xmax=1.1,
            ymin=0.0, ymax=0.6,
            xtick={0.3,0.7,1.1},
            ytick={0.1,0.2,0.3,0.4,0.5},
            legend pos=north east,
            grid=major,
            width=7.6cm, height=4.7cm,
            legend style={
                font=\scriptsize, 
                column sep=0.2ex, 
                legend cell align= right, 
                inner sep=0.2pt 
            },
        ]

        \addplot[color=blue, solid, thick, mark=*] coordinates {
            (0.3, 0.25)
            (0.7, 0.08)
            (1.1, 0.15)
        };
        \addlegendentry{24 Pretrained}

        \addplot[color=blue, dashed, thick, mark=o] coordinates {
            (0.3, 0.28)
            (0.7, 0.52)
            (1.1, 0.26)
        };
        \addlegendentry{24 Finetuned}

        \addplot[color=red, solid, thick, mark=square*] coordinates {
            (0.3, 0.25)
            (0.7, 0.40)
            (1.1, 0.28)
        };
        \addlegendentry{42 Pretrained}

        \addplot[color=red, dashed, thick, mark=square] coordinates {
            (0.3, 0.28)
            (0.7, 0.45)
            (1.1, 0.30)
        };
        \addlegendentry{42 Finetuned}

        \draw[dashed,gray] (axis cs:0.3,0) -- (axis cs:0.3,0.6);
        \draw[dashed,gray] (axis cs:0.7,0) -- (axis cs:0.7,0.6);
        \draw[dashed,gray] (axis cs:1.1,0) -- (axis cs:1.1,0.6);

    \end{axis}
    \end{tikzpicture}
     \caption{Success Rate metric for LLaMA 8B}
    \label{fig:temperature_vs_top10}
\end{wrapfigure}
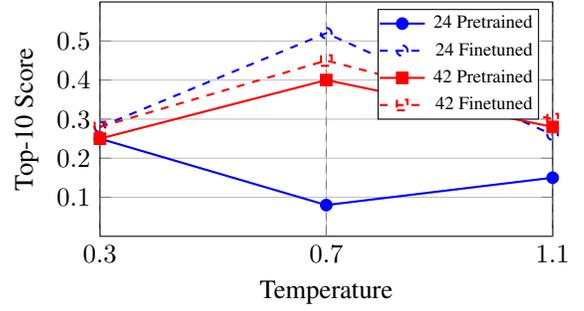

Our study explores the strengths and limitations of GFlowNets in generating diverse, accurate solutions and their transferability from the Game of 24 to the Game of 42 across varying temperature settings and decoding strategies. While fine-tuned GFlowNets outperform pre-trained LLaMA models on the Game of 24 tasks, their generalization to the Game of 42 is limited in accuracy (refer Figure \ref{fig:temperature_vs_top10}) and diversity. Computational limitations constrain these findings, as experiments focused on small and medium-sized LLMs. Expanding to larger models and diverse datasets could offer deeper insights into GFlowNet performance and transferability, with improving zero-shot transfer learning capabilities being key for broader task applicability.

\newpage 

\bibliography{iclr2025_conference}
\bibliographystyle{iclr2025_conference}

\appendix
\section{GFlowNets}
GFlowNets or GFNs \citep{bengio2023gflownet,bengio2021flow} were introduced as an alternative to traditional RL-based methods, trained to generate a final state $x$ with a sequence of steps (actions), sampled from a forward policy $P_F$ trained with an objective which makes it sample $x$ with probability proportional to reward $R(x)$ for that corresponding state $x$.

GFlowNets are preferred in applications where diversity of samples is desired but sampling is intractable and other exploration methods such as (MCMC) perform poorly. GFNs have been used for molecule generation tasks where action is incremental addition of simple building blocks and reward is estimated with the strength of binding the constructed molecule to a protein target. Unlike MCMC methods, GFNs do not suffer from the mixing problem, as each sample is generated independently.

\section*{Markovian Flow}

Let $G = (S, A)$ be a directed acyclic graph with the state space $S$ and action space $A$, $s_o \in S$ being the initial state of the graph. A specific state $s_t$ corresponds to some vertex, and an action corresponds to an edge $(s_t\rightarrow s_{t+1})$ in the graph.

A trajectory is defined as a sequence of transitions $\tau=(s_0 \rightarrow s_1\rightarrow \dots \rightarrow s_n)$ going from initial state $s_o$ to terminal state $s_n$. Let $\mathcal{T}$ be the set of all possible trajectories.

Define a non negative function $F:\mathcal{T} \rightarrow   \{0\} \cup \mathbb{R}^+$ called as the \textit{trajectory flow} function. From a flow perspective, the trajectory flow is equivalent to the amount of water flowing on a certain trajectory to reach the terminal state. Using this we define state flow as $F(s) = \sum_{s \in \tau} F(\tau)$ (all the water flowing through a state) and the edge flow as 
$$F(s \rightarrow s') = \sum_{\tau = (\dots \rightarrow s \rightarrow s' \rightarrow \dots)} F(\tau)$$
A non trivial flow $F$ defines a distribution over the set of possible trajectories $\mathcal{T}$ as 

$$P(\tau) = \frac{F(\tau)}{Z}, \;\;\;\text{where}\;\;\; Z=\sum_{\tau \in \mathcal{T}}F(\tau)$$

Using the Markovian assumption distribution $P$ can be factorized in terms of stepwise distributions $P_F(-|s)$ over the children of nonterminal states as follows

$$P(\tau = (s_0 \rightarrow \dots \rightarrow s_n)) = \prod_{t=1}^n P_F(s_t|s_{t-1})$$ 

similarly, there exist distributions $P_B(-|s)$ over the parents of each non initial state $s$ such that

$$P(\tau = (s_0 \rightarrow \dots \rightarrow s_n)|s_n) = \prod_{t=1}^n P_B(s_{t-1}|s_{t})$$ 

These distributions $P_F$ and $p_B$ are known as \textit{forward policy} and \textit{backward policy} respectively. Thus a forward policy can be used to sample terminal states by starting at the initial state $s_0$ and iteratively sampling actions from $P_F$ (equivalently for $P_B$). These can be computed in terms of edge flow and state flow as

$$P_F(s_t|s_{t-1}) = \frac{F(s_{t-1} \rightarrow s_t)} {F(s_{t-1})}, \; P_B(s_{t-1}|s_{t}) = \frac{F(s_{t-1} \rightarrow s_t)} {F(s_{t})}$$

Intuitively the forward policy distribution corresponds to how much fraction of water at state $s_{t-1}$ is flowing towards the state $s_{t}$, and backward policy distribution corresponds to how much fraction of water received at state $s_t$ is coming from the state $s_{t-1}$.

From the above equation, the \textit{detailed balance} constraint can be constructed.

$$F(s_{t-1})P_F(s_t|s_{t-1}) = F(s_t)P_B(s_{t-1}|s_t)$$

\section*{Training Objective}

 As discussed in the above sections, our aim for getting diverse trajectories was to learn a forward policy $P_F$ so that the resulting distribution over trajectories is proportional to the reward, i.e.

 $$F(s_n) = R(s_n) \;\;\; \forall \;\;\; \text{terminal state} \; s_n $$

To model such a forward policy, objectives such as detailed balance, trajectory balance, and subtrajectory balance are used \citep{bengio2023gflownet,bengio2021flow, madan2023learning, malkin2022trajectory}. \cite{yu2024flowreasoningtrainingllmsdivergent} uses trajectory balance objective as proposed in \citep{malkin2022trajectory}.

\section{LLM Reasoning}

\section*{Prompting methods}
Methods such as Chain-of-Thought (CoT)\citep{wei2022chain} prompt the model to guide its way through the intermediate reasoning steps to reach the final answer. Building on this idea, other tree and graph-based prompting methods such as ToT\citep{yao2024tree} and GoT\citep{besta2024graph}, have explored expanding the scope of reasoning by branching the intermediate steps into multiple paths. Despite their advantages, prompting methods depend highly on task-specific prompt design, which may limit their generalizability across domains.

\section*{Finetuning and RL Methods}
Fine-tuning strategies, such as Supervised Fine-Tuning (SFT) and Instruction Tuning (IT)  \citep{wei2021finetuned, mishra2021cross}, align LLM output with human preferences and task requirements by leveraging curated datasets. Although these methods improve accuracy and alignment, they often require significant effort to create high-quality labeled datasets, limiting scalability.
Reinforcement learning-based methods optimize models to maximize reward signals tied to desirable outputs, particularly Reinforcement Learning from Human Feedback (RLHF) \citep{ouyang2022training} and Proximal Policy Optimization (PPO) \citep{schulman2017proximal}. Although effective in enhancing alignment and precision, PPO and similar methods frequently prioritize generating high-reward solutions, which may inadvertently suppress output diversity.

\section{Examples}

\textbf{\large{Example for Game of 24:}} 

\textbf{Input: \( 2 \quad 4 \quad 8 \quad 10 \)}

Steps: 

8 / 4 = 2 (left: 2 10 2)

10 + 2 = 12 (left: 2 12)

2 * 12 = 24 (left: 24)

\textbf{Input: \( 2 \quad 4 \quad 8 \quad 10 \)}

Steps: 

2 + 10 = 12 (left: 4 8 12)

8 * 12 = 96 (left: 4 96)

96 / 4 = 24 (left: 24)

\textbf{Input: \( 2 \quad 4 \quad 8 \quad 10 \)}

Steps: 

2 + 10 = 12 (left: 4 8 12)

8 - 4 = 4 (left: 12 4)

12 - 4 = 8 (left: 8)

\textbf{Input: \( 2 \quad 4 \quad 8 \quad 10 \)}

Steps: 

2 + 10 = 12 (left: 4 8 12)

4 + 8 = 12 (left: 12 12)

12 + 12 = 24 (left: 24)

\textbf{Traj Num:  3}

\textbf{Result: SUCCESS}

\bigskip

\textbf{\large{Example for Game of 42:}} 

\textbf{Input: \( 2 \quad 12 \quad 3 \quad 6 \)}

Steps: 

2 + 3 = 5 (left: 12 6 5)

12 - 6 = 6 (left: 5 6)

5 + 6 = 11 (left: 11)

\textbf{Input: \( 2 \quad 12 \quad 3 \quad 6 \)}

Steps: 

2 * 12 = 24 (left: 3 6 24)

3 + 6 = 9 (left: 24 9)

24 - 9 = 15 (left: 15)

\textbf{Input: \( 2 \quad 12 \quad 3 \quad 6 \)}

Steps: 

2 * 6 = 12 (left: 12 3 12)

12 - 3 = 9 (left: 12 9)

9 + 12 = 21 (left: 21)

\textbf{Input: \( 2 \quad 12 \quad 3 \quad 6 \)}

Steps: 

2 * 12 = 24 (left: 3 6 24)

3 * 6 = 18 (left: 24 18)

24 + 18 = 42 (left: 42)

\textbf{Traj Num:  1}

\textbf{Result: Success}

\section{Results}
\label{res}
\begin{table*}[htbp]
\centering
\small
\begin{minipage}{0.48\textwidth}
\caption{SR of the models on GAME 24 Across Varying Top-K Values}
\vspace{5pt}
\centering
\begin{tabular}{lccc}
\hline
\textbf{Temperature} & \textbf{5} & \textbf{10} & \textbf{15} \\
\hline
\multicolumn{4}{c}{\textbf{LLaMA 1B}} \\
\hline
0.3 & 0.04 & 0.04 & 0.04 \\
0.7 & 0.12 & 0.06 & 0.08 \\
1.1 & 0.08 & 0.04 & 0.04 \\
\hline
\multicolumn{4}{c}{\textbf{LLaMA 1B Fine-Tuned with GFlowNet}} \\
\hline
0.3 & 0.26 & 0.26 & 0.26 \\
0.7 & 0.30 & 0.30 & 0.48 \\
1.1 & 0.26 & 0.20 & 0.20 \\
\hline
\multicolumn{4}{c}{\textbf{LLaMA 3B}} \\
\hline
0.3 & 0.12 & 0.12 & 0.12 \\
0.7 & 0.22 & 0.22 & 0.22 \\
1.1 & 0.08 & 0.12 & 0.10 \\
\hline
\multicolumn{4}{c}{\textbf{LLaMA 3B Fine-Tuned with GFlowNet}} \\
\hline
0.3 & 0.36 & 0.36 & 0.36 \\
0.7 & 0.48 & 0.40 & 0.40 \\
1.1 & 0.34 & 0.30 & 0.32 \\
\hline

\end{tabular}
\label{tab:game245}
\end{minipage}
\hfill
\begin{minipage}{0.48\textwidth}
\caption{SR of the models on GAME 42 Across Varying Top-K Values}
\vspace{5pt}
\centering
\begin{tabular}{lccc}
\hline
\textbf{Temperature} & \textbf{5} & \textbf{10} & \textbf{15} \\
\hline
\multicolumn{4}{c}{\textbf{LLaMA 1B}} \\
\hline
0.3 & 0.10 & 0.10 & 0.10 \\
0.7 & 0.12 & 0.12 & 0.20 \\
1.1 & 0.14 & 0.12 & 0.16 \\
\hline
\multicolumn{4}{c}{\textbf{LLaMA 1B Fine-Tuned with GFlowNet}} \\
\hline
0.3 & 0.10 & 0.10 & 0.10 \\
0.7 & 0.26 & 0.16 & 0.24 \\
1.1 & 0.14 & 0.16 & 0.20 \\
\hline
\multicolumn{4}{c}{\textbf{LLaMA 3B}} \\
\hline
0.3 & 0.22 & 0.22 & 0.22 \\
0.7 & 0.16 & 0.16 & 0.24 \\
1.1 & 0.14 & 0.14 & 0.22 \\
\hline
\multicolumn{4}{c}{\textbf{LLaMA 3B Fine-Tuned with GFlowNet}} \\
\hline
0.3 & 0.16 & 0.16 & 0.16 \\
0.7 & 0.20 & 0.20 & 0.20 \\
1.1 & 0.26 & 0.24 & 0.20 \\

\hline
\end{tabular}

\label{tab:game426}
\end{minipage}
\end{table*} 

\begin{table*}[htbp]
\centering
\small
\begin{minipage}{0.48\textwidth}
\caption{SR of the models on GAME 24 Across Varying Min-p Values}
\vspace{5pt}
\centering
\begin{tabular}{lccc}
\hline
\textbf{Temperature} & \textbf{0.05} & \textbf{0.10} & \textbf{0.15} \\
\hline
\multicolumn{4}{c}{\textbf{LLaMA 1B}} \\
\hline
0.3 & 0.04 & 0.04 & 0.02 \\
0.7 & 0.10 & 0.16 & 0.06 \\
1.1 & 0.06 & 0.08 & 0.04 \\
\hline
\multicolumn{4}{c}{\textbf{LLaMA 1B Fine-Tuned}} \\
\hline
0.3 & 0.22 & 0.28 & 0.28 \\
0.7 & 0.26 & 0.24 & 0.22 \\
1.1 & 0.20 & 0.16 & 0.26 \\
\hline
\multicolumn{4}{c}{\textbf{LLaMA 3B}} \\
\hline
0.3 & 0.12 & 0.12 & 0.10 \\
0.7 & 0.24 & 0.24 & 0.18 \\
1.1 & 0.22 & 0.20 & 0.18 \\
\hline
\multicolumn{4}{c}{\textbf{LLaMA 3B Fine-Tuned with GFlowNet}} \\
\hline
0.3 & 0.36 & 0.30 & 0.26 \\
0.7 & 0.46 & 0.38 & 0.30 \\
1.1 & 0.30 & 0.34 & 0.36 \\
\hline
\end{tabular}

\label{tab:game247}
\end{minipage}
\hfill
\begin{minipage}{0.48\textwidth}
\caption{SR of the models on GAME 42 Across Varying Min-p Values}
\vspace{5pt}
\centering
\begin{tabular}{lccc}
\hline
\textbf{Temperature} & \textbf{0.05} & \textbf{0.10} & \textbf{0.15} \\
\hline
\multicolumn{4}{c}{\textbf{LLaMA 1B}} \\
\hline
0.3 & 0.12 & 0.10 & 0.08 \\
0.7 & 0.18 & 0.18 & 0.22 \\
1.1 & 0.08 & 0.18 & 0.18 \\
\hline
\multicolumn{4}{c}{\textbf{LLaMA 1B Fine-Tuned}} \\
\hline
0.3 & 0.06 & 0.08 & 0.06 \\
0.7 & 0.20 & 0.16 & 0.16 \\
1.1 & 0.28 & 0.22 & 0.20 \\
\hline
\multicolumn{4}{c}{\textbf{LLaMA 3B}} \\
\hline
0.3 & 0.14 & 0.14 & 0.12 \\
0.7 & 0.28 & 0.22 & 0.16 \\
1.1 & 0.24 & 0.16 & 0.20 \\
\hline
\multicolumn{4}{c}{\textbf{LLaMA 3B Fine-Tuned with GFlowNet}} \\
\hline
0.3 & 0.10 & 0.12 & 0.12 \\
0.7 & 0.26 & 0.18 & 0.20 \\
1.1 & 0.12 & 0.20 & 0.21 \\
\hline

\end{tabular}

\label{tab:game428}
\end{minipage}
\end{table*} 

\begin{table*}[htbp]
\centering
\small
\begin{minipage}{0.48\textwidth}
\caption{SR of the models on GAME 24 Across Varying Top-p Values}
\vspace{5pt}
\centering
\begin{tabular}{lccc}
\hline
\textbf{Temperature} & \textbf{0.85} & \textbf{0.90} & \textbf{0.95} \\
\hline
\multicolumn{4}{c}{\textbf{LLaMA 1B}} \\
\hline
0.3 & 0.02 & 0.04 & 0.06 \\
0.7 & 0.10 & 0.14 & 0.12 \\
1.1 & 0.10 & 0.08 & 0.06 \\
\hline
\multicolumn{4}{c}{\textbf{LLaMA 1B Fine-Tuned with GFlowNet}} \\
\hline
0.3 & 0.22 & 0.36 & 0.20 \\
0.7 & 0.24 & 0.26 & 0.34 \\
1.1 & 0.24 & 0.20 & 0.36 \\
\hline
\multicolumn{4}{c}{\textbf{LLaMA 3B}} \\
\hline
0.3 & 0.06 & 0.12 & 0.12 \\
0.7 & 0.22 & 0.16 & 0.22 \\
1.1 & 0.22 & 0.12 & 0.16 \\
\hline
\multicolumn{4}{c}{\textbf{LLaMA 3B Fine-Tuned with GFlowNet}} \\
\hline
0.3 & 0.30 & 0.30 & 0.32 \\
0.7 & 0.32 & 0.36 & 0.40 \\
1.1 & 0.26 & 0.32 & 0.28 \\
\hline

\end{tabular}
\label{tab:game249}
\end{minipage}
\hfill
\begin{minipage}{0.48\textwidth}
\caption{SR of the models on GAME 42 Across Varying Top-p Values}
\vspace{5pt}
\centering
\begin{tabular}{lccc}
\hline
\textbf{Temperature} & \textbf{0.85} & \textbf{0.90} & \textbf{0.95} \\
\hline
\multicolumn{4}{c}{\textbf{LLaMA 1B}} \\
\hline
0.3 & 0.10 & 0.10 & 0.10 \\
0.7 & 0.22 & 0.18 & 0.16 \\
1.1 & 0.14 & 0.34 & 0.16 \\
\hline
\multicolumn{4}{c}{\textbf{LLaMA 1B Fine-Tuned with GFlowNet}} \\
\hline
0.3 & 0.04 & 0.06 & 0.08 \\
0.7 & 0.14 & 0.14 & 0.20 \\
1.1 & 0.32 & 0.36 & 0.20 \\
\hline
\multicolumn{4}{c}{\textbf{LLaMA 3B}} \\
\hline
0.3 & 0.14 & 0.20 & 0.16 \\
0.7 & 0.18 & 0.22 & 0.18 \\
1.1 & 0.24 & 0.20 & 0.20 \\
\hline
\multicolumn{4}{c}{\textbf{LLaMA 3B Fine-Tuned with GFlowNet}} \\
\hline
0.3 & 0.04 & 0.08 & 0.16 \\
0.7 & 0.30 & 0.14 & 0.20 \\
1.1 & 0.30 & 0.16 & 0.16 \\

\hline
\end{tabular}

\label{tab:game4210}
\end{minipage}
\end{table*} 

For diversity analysis

\begin{table*}[htbp]
\centering
\small
\begin{minipage}{0.48\textwidth}
\caption{SR of the models on GAME 24}
\vspace{5pt}
\centering
\begin{tabular}{lccc}
\hline
\textbf{Temperature} & \textbf{Top-10} & \textbf{Min-0.05} & \textbf{Top-0.85} \\
\hline
\multicolumn{4}{c}{\textbf{LLaMA 1B}} \\
\hline
0.3 & 0.20 & 0.20 & 0.20 \\
0.7 & 0.10 & 0.50 & 0.40 \\
1.1 & 0.50 & 0.50 & 0.30 \\
\hline
\multicolumn{4}{c}{\textbf{LLaMA 1B Fine-Tuned with GFlowNet}} \\
\hline
0.3 & 0.70 & 0.70 & 0.40 \\
0.7 & 0.30 & 0.80 & 0.60 \\
1.1 & 0.40 & 0.50 & 0.60 \\
\hline
\multicolumn{4}{c}{\textbf{LLaMA 3B}} \\
\hline
0.3 & 0.70 & 0.70 & 0.40 \\
0.7 & 0.50 & 0.60 & 0.50 \\
1.1 & 0.60 & 0.50 & 0.50 \\
\hline
\multicolumn{4}{c}{\textbf{LLaMA 3B Fine-Tuned with GFlowNet}} \\
\hline
0.3 & 0.70 & 0.50 & 0.70 \\
0.7 & 0.60 & 0.80 & 0.50 \\
1.1 & 0.70 & 0.70 & 0.80 \\

\hline
\multicolumn{4}{c}{\textbf{LLaMA 8B}} \\
\hline
0.3 & 0.60 & 0.45 & 0.45 \\
0.7 & 0.45 & 0.82 & 0.36 \\
1.1 & 0.30 & 0.55 & 0.55 \\
\hline
\multicolumn{4}{c}{\textbf{LLaMA 8B Fine-Tuned with GFlowNet}} \\
\hline
0.3 & 0.70 & 0.74 & 0.46 \\
0.7 & 0.70 & 0.65 & 0.90 \\
1.1 & 0.35 & 0.45 & 0.88 \\
\hline

\end{tabular}

\label{tab:game24-div}
\end{minipage}
\hfill
\begin{minipage}{0.48\textwidth}
\caption{SR of the models on GAME 42}
\vspace{5pt}
\centering
\begin{tabular}{lccc}
\hline
\textbf{Temperature} & \textbf{Top-10} & \textbf{Min-0.05} & \textbf{Top-0.85} \\
\hline
\multicolumn{4}{c}{\textbf{LLaMA 1B}} \\
\hline
0.3 & 0.00 & 0.00 & 0.05 \\
0.7 & 0.05 & 0.05 & 0.10 \\
1.1 & 0.10 & 0.10 & 0.10 \\
\hline
\multicolumn{4}{c}{\textbf{LLaMA 1B Fine-Tuned with GFlowNet}} \\
\hline
0.3 & 0.00 & 0.05 & 0.05 \\
0.7 & 0.20 & 0.20 & 0.20 \\
1.1 & 0.15 & 0.20 & 0.10 \\
\hline
\multicolumn{4}{c}{\textbf{LLaMA 3B}} \\
\hline
0.3 & 0.30 & 0.25 & 0.15 \\
0.7 & 0.40 & 0.15 & 0.40 \\
1.1 & 0.15 & 0.25 & 0.40 \\
\hline
\multicolumn{4}{c}{\textbf{LLaMA 3B Fine-Tuned with GFlowNet}} \\
\hline
0.3 & 0.10 & 0.10 & 0.10 \\
0.7 & 0.15 & 0.15 & 0.15 \\
1.1 & 0.25 & 0.20 & 0.10 \\
\hline
\multicolumn{4}{c}{\textbf{LLaMA 8B}} \\
\hline
0.3 & 0.25 & 0.18 & 0.35 \\
0.7 & 0.55 & 0.17 & 0.30 \\
1.1 & 0.27 & 0.21 & 0.25\\
\hline
\multicolumn{4}{c}{\textbf{LLaMA 8B Fine-Tuned with GFlowNet}} \\
\hline
0.3 & 0.24 & 0.30 & 0.30 \\
0.7 & 0.90 & 0.30 & 0.40 \\
1.1 & 0.45 & 0.15 & 0.35 \\
\hline
\end{tabular}

\label{tab:game42-div}
\end{minipage}
\end{table*}

\end{document}

%% file: math_commands.tex
\usepackage{amsmath,amsfonts,bm}

\def\eqref#1{equation~\ref{#1}}

\def\1{\bm{1}}

\DeclareMathAlphabet{\mathsfit}{\encodingdefault}{\sfdefault}{m}{sl}
\SetMathAlphabet{\mathsfit}{bold}{\encodingdefault}{\sfdefault}{bx}{n}

